\documentclass[12pt,fleqn,leqno]{article}  
\usepackage{latexsym}
\usepackage{graphics}   
\usepackage{epsf}
\newcommand{\cause}{\ensuremath{~causes~}}
\newcommand{\isa}{\ensuremath{\rightarrow_{IS-A}}}
\newcommand{\isauser}{\ensuremath{\rightarrow_{IS-A (object)}}}
\newcommand{\explicc}[2]{\ensuremath \;~\textit{explains}\;~#1\;~\textit{because\_possible}\;~#2 }
\newcommand{\explsf}[2]{\ensuremath \;\textit{expl}\;#1\;\textit{bec\_poss}\;#2 }
\newcommand{\explgf}[2]{\ensuremath \;\textit{explains}\;#1\;\textit{because\_possible}\;#2 }

\title{Using ASP with recent extensions for causal explanations%
\thanks{Paper presented at ASPOCP10, 
Answer Set Programming and Other Computing Paradigms Workshop, associated with ICLP, Edinburgh, July 20, 2010.}}
\author{Yves Moinard\vspace{1ex}\\
{\small INRIA Bretagne Atlantique, IRISA, Campus de Beaulieu,} \\ 
{\small 35042 Rennes cedex, France \hspace{1em}email: {moinard@irisa.fr}}}
\date{}

\begin{document}
\maketitle
\begin{abstract}
We examine the practicality for a user of using Answer Set Programming (ASP) for representing logical formalisms.
We choose as an example a formalism aiming at capturing causal explanations from causal 
information. We provide an implementation, 
showing the naturalness and relative efficiency of this translation job.
We are interested in the ease for writing an ASP program, in accordance with the claimed 
``declarative'' aspect of ASP. Limitations of the earlier systems (poor data structure and difficulty in reusing pieces of programs) made that in practice, the ``declarative aspect'' was more theoretical than practical. We show how recent improvements in working ASP systems
facilitate a lot the translation, even if 
a few improvements could still be useful.
\end{abstract}

\section{Introduction}
We examine a few difficulties encountered when trying to translate a logical formalism into
a running answer set programming (ASP) program, showing how recent developments in ASP systems are of great help.
We are not concerned here by complexity matters (also an important matter,
for which exists a large literature). Rather, we deal with the ease for writing, and more importantly reading (thus modifying easily) programs in ASP for a given problem.
Indeed, ASP is presented as (and is) a declarative formalism where it should be immediate to 
write a program from a logical formalization of a given problem. In practice, this is rarely 
as easy as claimed.
We choose as an example a formalism that we have designed in collaboration with Philippe Besnard
and Marie-Odile Cordier. This formalism aims at a logical formalization of explanations from causal and ``is-a'' statements. Given some information such as ``fire causes smoke'' and ``grey smoke is a smoke'', if ``grey smoke'' is established, we want to infer that ``fire'' is a (tentative) explanation for this fact.
The formalization \cite{BCM07,BCM08} is expressed in terms of {\em rules} 
such as

``if $\alpha \cause \beta$ and $\delta \  is a \ \beta$, then
$\alpha \ explains \ \delta \ because \ \{\alpha,\delta\} \ is\ possible$''.

Here, $\alpha,...$ may be first order atoms (without function symbols).
Thus, we can express these rules in a ``Datalog'' formulation. 
When various explanations are possible,   
some of them can be subsumed by other ones, and the set of the solutions should 
be pruned.
This concerns looking for paths in some graph and ASP is good for these tasks.
There currently exist systems which are rather efficient,
such as DLV\footnote{http://www.dbai.tuwien.ac.at/proj/dlv/} \cite{LPFEGPS06} 
or gringo/clasp or claspD\footnote{http://potassco.sourceforge.net/}.
Transforming formal rules into an ASP program is easy.
ASP should then be an interesting tool for researchers when designing a new theoretical
formalization as the one examined here.
An ASP program should help examining a great number of middle sized examples, 
and check whether the results are in accordance with our expectations. 
Then if middle sized examples could work, 
a few more optimization techniques could make real sized examples work. 
Since ASP rules are close to a natural language,
a final user could easily adapt a general framework 
to his precise needs, without requiring a complex specific system.

Even if ASP allows such direct and efficient translation, 
a few problems complicate the task.

Firstly, the poor data types available in pure ASP systems is a real drawback.
Our rules involve sets. There are many ways to represent sets in existing ASP systems. We had designed programs working in this way \cite{Moi07} 
(for systems not allowing functional terms), 
by representing a set as follows: $Expl(I,J,N,E)$ meaning there exists an ``explanation'' from $I$ to $J$ 
with a set of conditions which is the set $\{E / Expl(I,J,N,E)\}$, where $N$ is an 
index, necessary when $M$ sets of conditions exist ($N \in \{1,\cdots,M\}$).
However, the program is rather hard to read, and thus to be adapted.
Part of this difficulty comes from the second drawback given now.

Secondly, in pure ASP, it is hard to reuse part of a program. 
Similar rules should be written again, in a slightly different way.
Thus, e.g. the rules necessary to deal with sets should be rewritten in many parts of the full program. 

Thirdly, there are restrictions concerning ``brave'' or ``cautious'' reasoning.
In ASP, ``the problem is the program'' and the solution
consists in one or several sets of atoms satisfying the problem.
Each such set is an {\em answer set}. These answer sets are the ``ASP models''.
{\em Brave} (respectively {\em cautious}) solutions mean to look for atoms true 
in some answer set (respectively all the answer sets). 
In the existing ASP systems, this is generally possible, 
but in a restricted way only.

All these difficulties make that in practice, the claimed advantage 
in favour of the use of ASP is not so clear when the final program is written.
The rules in the program are encumbered by various tricks necessary to overcome
these limitations, and any subsequent modification in the program becomes complex.

However, things are evolving. To take the example of DLV, points 1 and 2  above are partly solved: DLV-Complex\footnote{http://www.mat.unical.it/dlv-complex} deals with the data structure problem, since it admits sets and lists. 
DLT \footnote{https://www.mat.unical.it/ianni/wiki/dlt} allows the use of ``templates'',
which solves to a great extent the problem for reusing part of a program. 
It is not yet possible to use these two improvements together. Since here the most problematic case was due to the use of sets, we have used DLV-Complex, without DLT. When DLT will be able to work with DLV-Complex, the task will be yet easier.

In the next section, we present what should be known about our explanation formalism
in order to understand its ASP translation.
Then, in section \ref{secASP}, we describe its ASP translation, explaining the interest, and the drawback, of using ASP for this kind of job.
Finally, we conclude by a few reasonable expectations about
the future ASP systems which could help a final user, and make ASP a really interesting
programming paradigm for such problems.

\section{The causal explanation formalism}

\subsection{Preliminaries}\label{subexplic}

For simplicity, we fully present the propositional version only.
The full formalism, with predicates (without functions) and with an elementary ``ontology" has been described in \cite{BCM07,BCM08}.
The extension from the propositional case to the general case 
is not difficult, as will be explained in \S \ref{subpred}.
We distinguish various types of statements in our formal system:
\begin{itemize}
\item[$C$:]
A theory expressing causal statements. \\
E.g. $On\_alarm \cause
H\!eard\_bell$ or $Flu \cause Fever\_Temperature$.
\item[$O$:]
An ontology in the form of a set of $IS$-$A$ links between two items
which can appear in a causal statement.\\ 
E.g., $Temperature\_39 \isa Fever\_Temperature$,\\ 
\makebox[2em]{} $Temperature\_41 \isa Fever\_Temperature$, \\
\makebox[2em]{} $Heard\_loud\_bell \isa Heard\_bell$,\\
\makebox[2em]{} $Heard\_soft\_bell \isa Heard\_bell$.
\item[$W$:]
A classical propositional theory expressing truths 
(incompatible facts, co-occurring facts, $\ldots$). E.g.,
$Heard\_soft\_bell \rightarrow \neg Heard\_loud\_bell$.
\end{itemize}
Intuitively, propositional symbols denote elementary properties
describing states of affairs, which can be {\em facts} or {\em events} 
such as $Fever\_Temperature$,\\ $On\_alarm$, $Heard\_bell$.
The causal statements express causal relations between facts or events
expressed by these propositional symbols.

Some care is necessary when providing these causal and ontological atoms.
If ``$Flu \cause Fever\_Temperature$'', we will conclude
{\em $Flu$ explains\\ $Temperature\_39$} from 
$Temperature\_39 \isa Fever\_Temperature$, but we cannot 
state $Flu \cause Temperature\_39$: we require that the causal information is
provided ``on the right level'' and in this case, $Temperature\_39$ is not 
on the right level, 
since ``$Temperature\_39$'' is too norrow
with respect to our knowledge about flu and temperatures. 

The formal system is meant to infer,
from such premises $C\cup O \cup W$, formulas denoting explanations.
This inference will be denoted $\vdash_C$. 
The ontological  atoms express some common sense knowledge which is necessary
to infer these  ``explanations''.
Notice that a feature of our formalism 
is that standard implication alone cannot help to infer explanations
\cite{BCM07,BCM08}.

In the following, $\alpha,\beta, \ldots$ denote the propositional 
atoms and $\Phi, \Psi, \ldots$ denote sets thereof. 
\begin{flushleft}\textbf{Atoms}\end{flushleft}
\begin{enumerate}
\item \emph{Propositional atoms}: $\alpha,\beta, \ldots$.
\item \emph{Causal atoms}: $\alpha \cause \beta$.
\item \emph{Ontological atoms}: $\alpha \isa \beta$.
\item \emph{Explanation atoms}: $\alpha \explicc{\beta}{\Phi}$.
\end{enumerate}
An ontological atom reads: $\alpha\mbox{ is a } \beta$.\\
An explanation atom reads: \emph{$\alpha$ is an explanation for $\beta$ because $\Phi$ is possible.}
\begin{flushleft}
\emph{Notation:} In order to help reading long formulas, explanation atoms are sometimes abbreviated as $\alpha \explsf{\beta}{\Phi}$.\\
\end{flushleft}
\textbf{Formulas}\nopagebreak
\begin{enumerate}
  \item \emph{Propositional formulas}: Boolean combinations of propositional
        atoms. 
  \item \emph{Causal formulas}: 
         Boolean combinations of causal or propositional atoms.
\end{enumerate}

The premises of the inference $\vdash_C$, namely $C \cup O \cup W$,
consist of propositional and causal formulas, and
ontological \emph{atoms} (no ontological formulas). Notice that explanation
atoms cannot occur in the premises.

The properties of causal and ontological formulas are as follows.
\begin{enumerate}
\item\label{propcausprop0} \textbf{Properties of the causal operator} 
  \begin{enumerate}
  \item \label{proofimplprop0} \emph{Entailing [standard] implication}:
      If $\alpha \cause \beta$, then $\alpha \rightarrow \beta.$
  \end{enumerate}
\item \label{propontprop0} \textbf{Properties of the ontological operator}
  \begin{enumerate}
  \item \label{ontoimplprop0} \emph{Entailing implication}:
      If $\alpha \isa \beta$, then $\alpha \rightarrow \beta.$
  \item\label{ontotransprop0} \emph{Transitivity}:
     If $a \isa b$ and $b \isa c$, then $a \isa c$.
  \item\label{ontorefprop0} \emph{Reflexivity}:
     $c \isa c$.
  \end{enumerate}
\end{enumerate}

Reflexivity is an unconventional property for an $IS$-$A$ hierarchy.
It is included here because it helps keeping the number of inference schemes low (note that in the ASP translation we do not need reflexivity).

$W$ is supposed to include (whether explicitly or via inference) all
the implications induced by the ontological atoms. 
For example, 
if\\
 $Heard\_loud\_bell \isa Heard\_bell$ is in $O$ then\\ 
$Heard\_loud\_bell \rightarrow Heard\_bell$ is in $W$.

Similarly, $W$ is supposed to include all conditionals induced by the causal
statements in $C$. For example,  
if $Flu \cause Fever\_Temperature$ is in $C$, then $Flu \rightarrow Fever\_Temperature$ is \mbox{in $W$}.

\subsection{The formal system }\label{proofsystem}

The above ideas are embedded in a short proof system extending classical logic:

\begin{enumerate}
\item\label{propcauschema0} 
\textbf{\emph{Causal atoms entail implication:}} 
     $(\alpha \cause \beta) \rightarrow (\alpha \rightarrow \beta)$.
\item\label{propontschema0} \textbf{\emph{Ontological atoms}}
  \begin{enumerate}
  \item\label{ontoimplicschema0} entail implication:
     If $\beta \isa \gamma$ then $\beta \rightarrow \gamma$.
  \item\label{ontotranschema0}transitivity: 
     If $\alpha \isa \beta$ and $\beta \isa \gamma$ then $\alpha \isa \gamma$.
  \item\label{ontorefschema0} reflexivity:
     $\alpha \isa \alpha$
  \end{enumerate}
\item\label{explicprop5} \textbf{\emph{Generating the explanation atoms}}
  \begin{enumerate}
 \item\label{explicontdnupprop5} {\em Initial case}\\
  $\begin{array}[t]{ll}
   \mbox{If} & \delta \isa  \beta, \;\;\delta \isa \gamma, \mbox{ and } \;
               W \not\models \neg (\alpha \wedge \delta),\\
   \mbox{then} \; & (\alpha \cause \beta) \quad \rightarrow \quad
        \alpha \explsf{\gamma}{\{\alpha,\delta\}}
      \end{array}$ 
  \item\label{transexplicprop5} {\em Transitivity (gathering the
      conditions)}\hspace{1em}
      If\hspace{1em} $W \not \models \neg \bigwedge (\Phi \cup\Psi)$,\\ then \hspace{1em}
       $(\alpha \explsf{\beta}{\Phi} \;\;\wedge \;\;  
                  \beta \explsf{\gamma}{\Psi})\\  
         \makebox[3.5em]{}\rightarrow \;\; \alpha \explsf{\gamma}{(\Phi \cup \Psi)}.$
  \item\label{simplexplicprop5} {\em Simplification of the set of conditions}\hfill
      If \hfill 
        $W \models \bigwedge\Phi \rightarrow \bigvee_{i=1}^n \bigwedge\Phi_i$,\\
        then \hspace{.2em}
       $\bigwedge_{i \in \{1,\cdots,n\}}
            \alpha \explsf{\beta}{(\Phi_i \cup \Phi)}$
        \makebox[.3em]{}   $\rightarrow\hspace{.3em}
          \alpha \explsf{\beta}{\Phi}.$
  \end{enumerate}
\end{enumerate}

These schemes allow us to obtain the inference patterns evoked in the
previous section:

\noindent The most elementary ``initial case''
applies (\ref{ontorefschema0}) upon
(\ref{explicontdnupprop5}) where $\beta=\gamma=\delta$,
together with an obvious simplification (\ref{simplexplicprop5}) 
since $\alpha \rightarrow \beta$ here, getting:

\noindent If $\alpha \cause  \beta$ and  $W \not\models \neg \alpha$ then 
 $\alpha \explsf{\beta}{\{\alpha\}}$.

Two other particular cases read as follows (respectively 
$\gamma = \delta$ and $\beta = \delta$):

\noindent If $\alpha \cause \beta$, $\;\;\delta \isa  \beta$ and 
          $W \not\models \neg (\alpha \wedge \delta)$
   then  $\alpha \explsf{\delta}{\{\alpha,\delta\}}$.

\noindent If $\alpha \cause \beta$,  $\;\;\beta \isa \gamma$ and 
               $W \not\models \neg \alpha$
   then $\alpha \explsf{\gamma}{\{\alpha\}}$.\\
  
Notice that we do not allow explanation through the opposite ontological sequence
$\beta \isa \gamma$ and $\delta \isa \gamma$ [cf the general formulation \ref{explicontdnupprop5}].
This is from experiences with situations of the kind of those evoked in the preliminaries, 
and it is in accordance with the notion of ``right level''.
Suppose we would admit these sequences, then if e.g.\\
$\alpha \cause Heard\_loud\_bell$, we would necessarily get the conclusion\\
$\alpha \explsf{Heard\_soft\_bell}{\{\alpha, Heard\_soft\_bell\}}$,
from the natural data \\
$Heard\_loud\_bell \isa Heard\_bell$ and $Heard\_soft\_bell \isa Heard\_bell$.\\ 
We consider such a conclusion as unwanted in this situation.

It is possible that some  domains of application would need other rules.
Our rules are intended as a compromise between expressive power, naturalness of description and relatively efficient  computability.\\

The system is independent of any opinion about the controversial discussion about transitivity of {\em causation}.
However, we provide some transitivity for {\em explanations}
(the conditions are gathered then, cf \ref{transexplicprop5}).\\

The simplification rule (\ref{simplexplicprop5})
is powerful
and can be considered as of little interest with respect to the extra computational cost associated with it.
In practice, simpler rules could suffice, and our ASP translation implements
a weaker rule defined as follows:

[\ref{simplexplicprop5}'] 
If $W \models \bigwedge\Phi - \{\varphi\}  \rightarrow \bigwedge\Phi$,
        and $\alpha \explgf{\beta}{\Phi}$\\\makebox[3.6em]{}  then
          $\alpha \explgf{\beta}{\Phi- \{\varphi\} }.$

Notice a minor point: we consider as important to keep $\alpha$ is such a case,
thus the ASP translation does not apply this simplification (removing 
a redundant $\varphi$ from the set of conditions) when $\varphi=\alpha$.\\

It is important to introduce the notion of {\em optimal explanation atoms}:
An atom $\alpha \explgf{\beta}{\Phi}$ is {\em optimal} if there is no 
explanation atom
$\alpha \explgf{\beta}{\Psi}$ where $W\models \bigwedge\Phi \rightarrow \bigwedge\Psi$ while $W \not \models \bigwedge\Psi \rightarrow \bigwedge\Phi$.
Keeping only these weakest sets of conditions
is particularly useful when the derivation is made only thanks to the part of $W$ coming from Points \ref{propcauschema0} and \ref{ontoimplicschema0} above.
Indeed, this makes the result easier to read in case where many possible explanation atoms 
exist from $\alpha$ to $\beta$.
In this case, we are certain to keep all the relevant explanation atoms, 
even if some additional $W$ is introduced afterwards.

\subsection{A generic diagram}\label{subdiagram}

\begin{figure}[!b]
\includegraphics{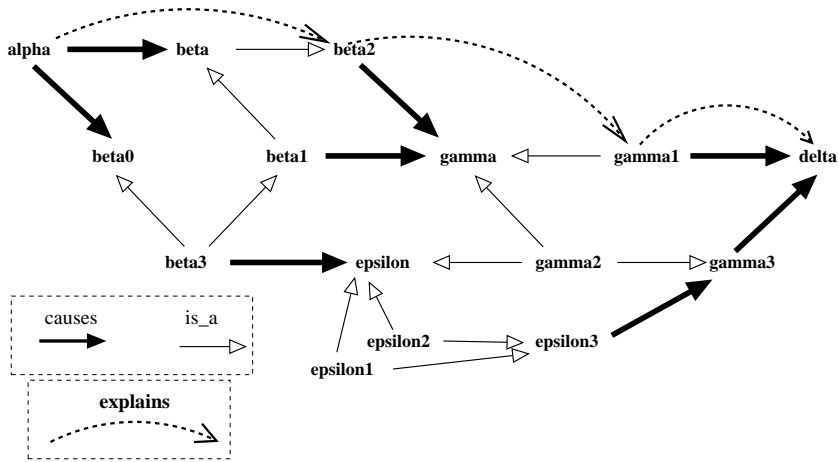}
\caption{A generic diagram, the theory with a first explaining path}  
\label{fig1}  \end{figure}

Below an abstract diagram is depicted
that summarizes many patterns of inferred explanations from various cases of
causal statements and $\isa$ links.
The theory is described as follows (see Figure~\ref{fig1}, with greek letters in their latin names):

\noindent$\alpha \cause \beta$, \hfill 
$\alpha \cause \beta_0$, \hfill 
$\beta_2 \cause \gamma$, \hfill 
$\beta_1 \cause \gamma$, \hfill $\,$\\
$\beta_3 \cause \epsilon$, \hfill 
$\gamma_1 \cause \delta$,  \hfill 
$\gamma_3 \cause \delta$, \hfill $\epsilon_3 \cause \gamma_3$; \hfill $\,$\\ 
$\beta \isa \beta_2$, \hfill $\beta_1 \isa \beta$, \hfill  
$\beta_3 \isa \beta_0$, \hfill $\beta_3 \isa \beta_1$, \hfill $\,$\\
$\gamma_1 \isa \gamma$, \hfill $\gamma_2 \isa \gamma$, \hfill 
$\gamma_2 \isa \gamma_3$, \hfill $\gamma_2 \isa \epsilon$, \hfill $\,$\\
$\epsilon_1 \isa \epsilon$, \hfill $\epsilon_2 \isa \epsilon$, \hfill 
$\epsilon_1 \isa \epsilon_3$, \hfill $\epsilon_2\isa \epsilon_3$. \hfill $\,$

\begin{figure}[h]
\includegraphics{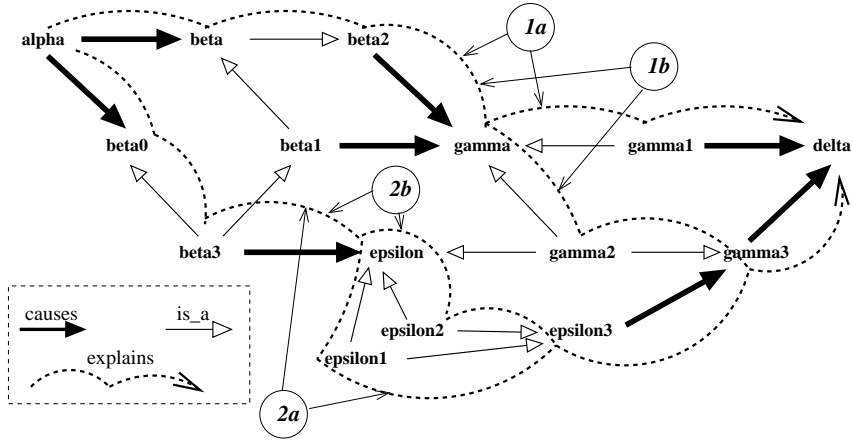}
\caption{Four optimal explanation paths from ``alpha'' to ``delta''}  %
\label{fig2} 
\end{figure}

This example shows various different ``explaining paths'' from a few given
causal and ontological atoms. 
Here there 
is a first ``explaining path'' from $\alpha$ to $\delta$
(Figure~\ref{fig1}, see also path $(1a)$ on Figure~\ref{fig2}). 
We get successively:\\
$\alpha \explsf{\beta_2}{\{\alpha\}}$,\hfill 
$\alpha \explsf{\gamma_1}{\{\alpha,\gamma_1\}}$,
\hfill and \hfill $\alpha \explsf{\delta}{\{\alpha,\gamma_1\}}.$\hfill $\,$

\noindent As another ``explaining path'', we get:\hfill
$\alpha \explsf{\delta}{\{\alpha,\beta_1,\gamma_1\}}.$
\hfill \makebox[0mm]{}

This second path is not optimal since
$\{\alpha,\gamma_1\} \subset \{\alpha,\beta_1,\gamma_1\}$.
The simplifying rule produces $\alpha\explsf{\delta}{\{\alpha,\gamma_1\}}$ 
from\\
$\alpha \explsf{\delta}{\{\alpha,\beta_1,\gamma_1\}}$ but, from a
computational point of view, it can be better not to generate the second path 
at all.

Here are the four optimal explanation atoms from $\alpha$ to $\delta$
(Fig.~\ref{fig2}):

\noindent$\;\;\begin{array}[t]{lclc}%
(1a)\;&\;\alpha \explsf{\delta}{\{\alpha, \gamma_1\}}\;\;& \;
(1b)\; &\;\alpha \explsf{\delta}{\{\alpha, \gamma_2\}}\\
(2a)\;&\;\alpha \explsf{\delta}{\{\alpha, \beta_3, \epsilon_1\}}\;\;& \;
(2b)\;&\;\alpha \explsf{\delta}{\{\alpha, \beta_3, \epsilon_2\}}.  
\end{array}$\\

\section{An ASP translation of the formalism}\label{secASP}

\subsection{Presentation}
We have implemented a program in DLV \cite{LPFEGPS06},
an implementation of the Answer Set Programming (ASP) formalism \cite{Bar03},
that takes only a few seconds to give all the results $\sigma1 \explsf{\sigma2}{\Phi}$,
for all examples of the kind depicted in
Fig.~\ref{fig2}. 
The first version \cite{Moi07} used pure DLV, and was encumbered with 
tricks allowing to deal with sets.
Now that DLV-Complex exists, the new version is simpler.
As already written, it does not make full ``simplification", however it makes the most important ones.
For greater examples, it still works, if we omit the 
simplification/optimization step.
We have tried a much greater example, involving two different copies of the example
of the diagram, linked through an additional small set of data.
This ends up with an example with more than a hundred symbols and more than 10 
different explanation atoms for some $(I,J)$.
This is not a ``real world example'' yet, but it is not too small, and it shows that we are 
close to realistic examples.
Hopefully, the progress in ASP systems will make that in a near future real world examples could work.\\

As explained below, we have encountered a fourth problem, not listed in the ``three problems" evoked above.
Indeed, for this ``big example", the full program, including simplification and testing the set of conditions, did not work on our computer
(crash after more than one hour...).
However, since the simplification/optimization step is clearly separated from 
the first generating step, and since the verification is separated also, we have separated the program into  three successive programs:\\

The first one generates explanation atoms (not all of them, but sufficiently 
many to  be able to retrieve all the optimal ones from those produced here).\\

Then, starting from the results of this first program, a second program makes all the relevant simplifications and optimizations, in order to help reading the 
set of the solutions. The simplification does not take disjunction into 
account as in the powerful 
rule in Point \ref{simplexplicprop5} in \S \ref{proofsystem}, rather it 
corresponds to Point \ref{simplexplicprop5}'.
When two possible sets $\Phi$ and $\Psi$ of conditions exist for some $(I,J)$,
if each $\psi \in\Psi$ is entailed (defined as above) by some $\varphi \in \Phi$,
and if the contrary is not true, then the explanation atom
$I\explsf{J}{\Phi}$ is disregarded, only the explanation atom with the weaker condition set $I\explsf{J}{\Psi}$ (the most likely to be satisfied)
is kept as a result.
This tends to keep only optimal explanation atoms defined at the end 
of \S \ref{proofsystem} above, while remaining efficient enough
(similarly to the simplification part, only a reasonably efficient part of the 
 formal definition is implemented).
This facilitates the human reading of the result, and in fact it is almost mandatory with some sets of data.\\

A third program takes the result of the second one (or directly of the first one
if the second seems useless) and checks whether the set of conditions is satisfied in the answer set considered.
In fact, in the formalism, we should check if it is true in some answer set.
This needs full brave reasoning, which for now is not possible with the running systems. However, in our separated programs (not detailed here), it is easier to
regroup all the answer sets into a unique big one, where the original answer sets are distinguished by an index (from 1 to the number of original answer sets).
In this way, we can do brave and cautious reasoning,
and check the set of conditions as described in the formalism.\\

This exhibits a fourth problem with the existing systems.
This problem has been addressed in various papers about ASP, but apparently it has not yet given rise to a real running system.
Indeed, \cite{TBA05} describes a system which allows to enumerate the answer sets, which is what we need here. However, no running system is referenced in this very interesting paper, which describes small, natural and very useful improvements for ASP systems.
A more recent paper \cite{BDT06} describes a more ambitious system which also
deals with this point.
Hopefully, these very interesting and natural improvements will be introduced in available 
systems in a near future.

\subsection{The generating part: getting the relevant explanation atoms}\label{subgen}

The ``answer'' of an ASP program is a set of {\em answer sets}, that is a set of 
concrete literals satisfying the rules (see e. g. \cite{Bar03} for exact definitions
and \cite{LPFEGPS06} for the precise syntax of DLV).

The user provides the following data:

\noindent 
{\sf symbol(alpha).}\hspace{1em} for each propositional symbol {\sf alpha}
[mandatory only if the symbol does not appear in a causal, ontological or classical atom].\\
 {\sf cause(alpha,beta).} \hspace{1em} for each causal atom $\alpha \cause \beta$,\\
{\sf ont(alpha,beta).} \hspace{1em} for each ontological atom $\alpha \isa \beta$,\\
{\sf true(alpha).}  or  {\sf -true(alpha).} \hspace{.3em} for each propositional atom $\alpha$ true or false.

Causal and propositional formulas must be put in conjunctive normal form, in order to be entered as sets of clauses such as

\noindent{\sf \{-true(epsilon1) v -true(gamma1) v -true(gamma2).}\\
{\sf -true(epsilon2) v -true(gamma1) v -true(gamma2).\}} 

\noindent for the formula
$ (\neg\epsilon1 \wedge \neg \epsilon2) \vee \neg \gamma1 \vee \neg \gamma2$;

\noindent or
{\sf \{cause(beta2,gamma) v cause(epsilon3,gamma3).\}} 

\noindent for
$(\beta2 \cause \gamma) \vee (\epsilon3 \cause \gamma3).$.

Notice that if we really need all the logical models, at least with respect to some
propositional or causal atoms, we must ``complete'' each
propositional or causal atom concerned as follows:

{\sf true(alpha) v -true(alpha).} or 
{\sf cause(alpha,beta) v -cause(alpha,beta).}

This is generally not necessary, and should be avoided as far as possible
since it is computationally demanding.

The interesting result consists in the explanation predicates:

\noindent {\sf ecSet(alpha,beta,\{alpha,delta,gamma\})} represents the explanation atom

\noindent $\alpha \explsf{\beta}{\{\alpha,\delta,\gamma\}}$.

Here come the first rules:

\noindent {\sf ontt(I,J) :- ont(I,J). \hspace{1em} ontt(I,K) :- ontt(I,J), ont(J,K).} 
(cf \ref{ontotranschema0} \S \ref{proofsystem}).

For the sake of ``safety'' of some rules, and for defining {\sf impCO} 
introduced below we may need to define all the symbols, 
and the ones which can appear in an explanation set (suffix ``{\sf E}'').

\noindent  {\sf symbolE(X) :- cause(X,Y). \hspace{1em} 
symbolE(Y) :- cause(X,Y).\\
symbolE(X) :- ont(X,Y). \hspace{1em} 
symbolE(Y) :- ont(X,Y).\\
symbol(X) :- symbolE(X).}

Implication derived from causal and ontological atoms (``s'' for ``strict''):

\noindent{\sf 
\begin{tabular}{rclrcl}%
impCO(I,J)&$\!\!\!$:-$\!\!$& cause(I,J). & impCO(I,J)$\!\!$&:-$\!\!$& ont(I,J).\\
impCO(I,K)&$\!\!\!$:-$\!\!$& impCO(I,J), impCO(J,K). &
impCO(I,I)$\!\!$&:-$\!\!$& symbolE(I).\\ 
impCOs(I,J)&$\!\!\!$:-$\!\!$& impCO(I,J), not impCO(J,I).
\end{tabular}}

We split the general basic generation rule 
\ref{explicontdnupprop5} \S \ref{proofsystem}, 
in order to improve the computational performances.
Indeed, in the first three particular cases, only one optimal initial explanation atom in {\sf (I,J)} exists, thus the computation can be simplified.

{\sf ecinit(I,J,E)} represents $I \explsf{J}{\{I,E\}}$ where this explanation atom is obtained without using the transitivity rule:

\noindent {\sf ecinit(I,J,I) :- cause(I,J). \hspace{1em}  
ecinit(I,J,I) :- cause(I,X), ontt(J,X), impCO(I,J). \\
ecinit(I,J,I) :- cause(I,X), ontt(X,J). \\  
ecinit(I,J,I) :- cause(I,X), ontt(E,X), ontt(E,J), impCO(I,E).}

The most complicated case (with the two ontological axioms) may lead to several
explanation atoms from $I$ to $J$, which requires some complications:

\noindent {\sf ecinit3p(I,J,E) :- ecinit(I,E,E), cause(I,X), ontt(E,X), ontt(E,J), 
                   not ecinit(I,J,I),\\\indent not ecinit(I,J,J).

\noindent nonecinit(I,J,E) :- ecinit3p(I,J,E1), ecinit3p(I,J,E), impCOs(E,E1).}

\noindent {\sf ecinit(I,J,E) :- ecinit3p(I,J,E), not nonecinit(I,J,E).}\\
\indent[Avoids keeping clearly non optimal ones.]

Since the set of conditions are singletons or pairs in the initial explanation 
atoms [{\sf ecinit}], set representation was not required.
Real explanation atoms [{\sf ecSet}] are introduced now, together with 
explicit sets which allow a serious simplification of the ASP rules.
 
Firstly, we initialize {\sf ecSet} with {\sf ecinit}
[{\sf \#insert(Set1,E,Set)} means in DLV-Complex: $Set = Set1 \cup \{E\}$]:

\noindent {\sf ecSet(I,J,\{I\}) :- ecinit(I,J,I).\\
ecSet(I,J,\{I,J\}) :- ecinit(I,J,J), not ecSet(I,J,\{I\}).\\
ecSet(I,J,\{I,E\}) :- ecinit(I,J,E), not ecSet(I,J,\{I\}), not ecSet(I,J,\{I,J\}).}

Then comes the translation of Point \ref{transexplicprop5} \S \ref{proofsystem}.

\noindent {\sf ecSet(I,J,Set) :- ecSet(I,K,Set1), not ecSet(I,J,Set1), ecinit(K,J,E2), E2 != K,\\ \indent\#insert(Set1,E2,Set).

\noindent ecSet(I,J,Set) :- ecSet(I,K,Set), ecinit(K,J,K).}

[Nothing to add in this case, since we know that {\sf Set} $\vdash_C$ {\sf K}.]

And this is enough for getting all the relevant explanation atoms.
Only a few computational optimization tricks complicate a little bit the writing,
however, the ASP rules remain very close to the formal rules given in \S \ref{proofsystem}.

\subsection{Optimizing the explanation atoms}\label{subopt}

This is the most computationally demanding part.
Rigorously, it is not mandatory.
However, it is important in order to avoid providing too many 
unnecessary explanation atoms which complicate the interpretation of the result.

As a short example, suppose we have following data:

\noindent $\alpha \cause \beta$, \hfill $\alpha \cause \beta_0$, \hfill 
$\beta_2 \cause \gamma$, \hfill 
$\beta_1 \cause \gamma$, \hfill $\,$\\
$\beta_2 \isa \beta_0$, \hspace{2em} $\beta_1 \isa \beta$, 
\hspace{2em} $\beta_2 \isa \beta_1$.\\

Then, we get the following explanation atoms concerning $(\alpha,\gamma)$:

Expl1: $\alpha \explsf{\gamma}{\{\alpha,\beta_1\}}$ and 
Expl2: $\alpha \explsf{\gamma}{\{\alpha,\beta_2\}}$.

Since we have $\beta_2 \isa \beta_1$, we get $\beta_2 \rightarrow \beta_1$
from \ref{ontoimplicschema0} \S \ref{proofsystem}
in $W$, even if the user does not provide any explicit $W$.
Thus, each element in ${\{\alpha,\beta_1\}}$ is entailed by some element in 
${\{\alpha,\beta_2\}}$ and not conversely. Thus, the weaker set 
${\{\alpha,\beta_1\}}$ is more likely to be satisfied, whatever are the other data,
and in particular whatever may be an explicit $W$ given as additional data.
Thus, it is useless, and disturbing, to provide Expl2: Expl1 is enough.
It happens here that Expl1 is {\em optimal} with the given data.
Obviously, in more complex cases involving disjunction, the ``element wise'' test
would not be enough to discard all the sets of conditions which are too strong for entailment. However, the test described here is a good compromise between 
efficient computation and readability of the result.

We have tried various other programs providing only the [quasi] optimal sets,
in order to avoid this ``optimizing'' part.
All of them were much slower: it is better to provide first the explanation atoms with a program such as the one given in \S \ref{subgen}, which does not take great care for avoiding superfluous answers, and then to prune the set of solutions.

Here comes the ``optimizing part''.
Remind that not all the simplifications or optimizations possible are made.
As explained above, only those which are easy to compute are made. 
Indeed, except in artificially complicated data, most of the simplifications 
are made in this way, while dealing with the tricky cases would make the program too slow for a marginal advantage. These simplification/optimizations are mandatory in order to facilitate human reading, and could be omitted 
in  a purely formal perspective since anyway the produced explanation atoms 
cover all the possible situations.
The rules are very simple: 

{\sf imp(I,J) :- impCO(I,J).} (additional predicate {\sf imp} useless in the version presented here). 

Propagating the truth values:

\noindent {\sf true(J) :- true(I), imp(I,J). \hspace{1em}
-true(I) :- -true(J), imp(I,J).}

Eliminating sets of conditions which contain another set\\
\noindent[{\sf \#subSet(Set1,Set)} means $Set1 \subseteq Set$ and {\sf \#member(E,Set)} means $E \in Set$]:

\noindent {\sf toolargeSet(I,J,Set) :- ecSet(I,J,Set1), ecSet(I,J,Set), Set1 != Set,\\
 \indent\#subSet(Set1,Set).\\ 
ecSetsmall(I,J,Set) :- ecSet(I,J,Set), not toolargeSet(I,J,Set).}

\noindent {\sf impCOSetEl(Set1,E2) :- ecSetsmall(I,J,Set1), ecSetsmall(I,J,Set2), \\
\indent Set1 != Set2, impCO(E1,E2), \#member(E1,Set1), \#member(E2,Set2),\\
\indent not \#member(E1,Set2), not \#member(E2,Set1). } 

\noindent {\sf nonImpCOSet(Set1,Set2) :- ecSetsmall(I,J,Set1), ecSetsmall(I,J,Set2),\\
\indent  Set1 != Set2, symbolE(E2), \#member(E2,Set2),not \#member(E2,Set1),\\
\indent   not impCOSetEl(Set1,E2).}\hspace{1em}
\% ( ``{\sf not \#member(E2,Set1),}'' is optional) 

\noindent {\sf toostrongSet(I,J,Set) :- ecSetsmall(I,J,Set), ecSetsmall(I,J,Set1),\\
\indent  Set != Set1,  nonImpCOSet(Set1,Set), not nonImpCOSet(Set,Set1).}

\noindent {\sf ecSetRes(I,J,Set) :- ecSetsmall(I,J,Set), not toostrongSet(I,J,Set).} 

As a result, we only keep the explanation atoms {\sf ecSetRes(I,J,Set)} 
where no element can be removed from {\sf Set} and where no set {\sf StrongSet}
is such that {\sf ecSet(I,J,StrongSet)} and each element of {\sf Set} is implied 
(in the meaning of {\sf impCO}) by some element of {\sf StrongSet}, and not conversely.

This program is rather slow:
the two programs (\S \ref{subgen} and the present \ref{subopt})
can be launched together for examples such as in the diagram of \S \ref{subdiagram}
and for slightly larger example, but it is impossible for our ``big example''.
However, it is possible to launch the first program (instantaneous on our examples), and then the second one starting  with the results of the first one.
Then, our ``big example'' is solved in far less than one minute on our computer.

\subsection{Checking the set of conditions}\label{subckeck}

Finally, the following program starts from the result
of the preceding programs and checks, in each answer set, whether the set of conditions is satisfied or not. 
This program could also be launched starting from the result of the first program
(replacing {\sf ecSetRes} with {\sf ecSet}), 
simply superfluous explanation atoms would be checked also, and almost no
``optimization'' would be done.

The result is given by {\sf explVer(I,J,Set)}: $I \explsf{J}{Set}$ where
$Set$ is satisfiable in the answer set considered (``{\sf Ver}'' stands for ``verified'').

\noindent {\sf explSuppr(I,J,Set) :- ecSetRes(I,J,Set), -true(E), \#member(E,Set). 

\noindent explVer(I,J,Set) :- ecSetRes(I,J,Set), not explSuppr(I,J,Set).}

Notice that only ``individual'' checking is made here, in accordance with the requirement that the computational properties remain manageable.
In each answer set, this check is enough.

Notice that, as already evoked, in our running split program (not provided in full here for lack of space), all the answer sets are put into a single one, 
with an index parameter added for each old answer set. This allows to make real 
cautious reasoning when checking the consistency of the explanation sets
(and only for this point), in accordance with the formalism. 
Then, a few more rules could be added in order to get results in full
accordance with our formalism
(since e.g. answer sets do not provide all the classical models). However, even if these rules are written with care, it seems unlikely that the resulting program can run in practice for  great examples.
As already evoked above, if we want to get all the classical models,
a possibility is to require the {\em completion} of all the atoms 
(propositional or causal) which can provoke the 
existence of various answer sets, by adding the disjunctive rules

{\sf true(alpha) v -true(alpha).} and  
{\sf cause(gamma,delta) v -cause(gamma,delta).}\\
for each of these atoms involved in a non atomic formula.
This solution is easy to write in the program, but becomes clearly 
unmanageable from a computational point of view since the number of answer sets may explode. It is the standard, but not practical, way for getting all the classical models in ASP. Then, together with real cautious reasoning, checking for possible consistency of the sets could be made in full accordance with the formalism. Notice that it is not always clear whether the intended meaning 
is better rendered by classical models than by answer sets.\\

We have taken care to describe all the practical limitations of our solution.
However, apart from these (generally minor in practice) points, 
we hope to have convinced that ASP 
is very interesting in order to deal with this kind of problem.

One of the advantages is that if we want to modify a rule of the formalism, this can be done easily, since there is a close relation between the formal rules and the ASP rules.
The main difference between formal rules and the ASP rules described above 
concern the ``initial rule'' in the first program \S \ref{subgen}.
This difference comes from the fact that we have done our best to provide a program running in a reasonable time. This is another interest of using ASP: such computational optimizations can be introduced in a relatively natural way.
Even with these complications, modifying the rules remains easy.

Also, for the example described here, the gain of using DLV-Complex instead of 
pure DLV (or gringo/claspD) is significant and worth mentioning.

\subsection{Dealing with predicate logic}\label{subpred}
For the sake of conciseness we have given the translation of the propositional version of our formalism. In fact, this is not a serious concern. Indeed, the computational efficiency is similar, since the real tricky points all appear in the propositional version. A standard way to deal with predicates in ASP, starting from a propositional version, is to add a few parameters representing 
the predicate symbols.
As an example, let us consider atoms such as $heard(bell)$ or $like(bell)$ or 
$own(student,book)$ in place of $\alpha$ or $\beta$. In the formalism, all we need to do is to replace e. g. facts such as

\noindent $heard\_bell \cause on\_alarm.$ \hfill by \hfill
$heard(bell) \cause on(alarm).$,\\ 
and to replace all the rules accordingly.

In this way, we can also translate the slightly extended ontology described in \cite{BCM08}. There, two kinds of parameters are considered for each predicate:
the ones for which the predicate is {\em essentially universal} and the ones for which the predicate is {\em essentially existential}.
Let us suppose that we intend that $heard$ is essentially existential with respect to its parameter and $like$ essentially universal.
Intuitively, this means that $heard$ is intended as meaning {\em heard some}
while $like$ means {\em like all}.
The ontological information would be provided by the user as follows:
\hspace{1em} $loud\_bell \isauser bell$ \hspace{.5em} and \hspace{.5em} 
$white\_car \isauser car$.

This would provide the following $\isa$ relation between atoms

\noindent $heard(bell) \isa heard(loud\_bell).$ \hspace{1em} and \\
\noindent $like(car) \isa like(white\_car).$.\\

This means that the ASP formulation would contain the following facts and rules:
\hspace{1em} {\sf ont\_object(loud\_bell,bell)}.\hspace{1em}
{\sf ont\_object(white\_car,car)}.\\
\noindent {\sf ont([P,X],[P,Y]) :- onekind(P), ont\_object(X,Y).} \hspace{1em} 
and\\
\noindent {\sf ont([P,Y],[P,X]) :- allkind(P), ont\_object(X,Y).}\\

Dealing with lists (denoted inside brackets) of DLV-Complex is convenient: 
a small set of ASP rules can deal with 
predicates of any arity, even if we do not describe the full formalism here
for the sake of conciseness.

Propositional parameters would be dealt with the following rule:

\noindent {\sf ont([A],[B]) :- ont\_object(A,B), propkind(A), propkind(B).} 

Let us consider a  binary predicate $own$, with
$own(student,book)$ meaning that the individual (or object) $student$ 
(whatever it denotes in our formalism) owns the object called $book$.
Here, $own(X,Y)$ is intended to mean: ``{\em all X own some Y}''.
The following rule {\tt(O-O)} can deal with this predicate:

\noindent {\sf ont([P,X,Y],[P,X1,Y1]) :- all\_onekind(P), ont\_object(X1,X), ont\_object(Y,Y1).}

The user should state the kind of parameters for each predicate, as follows:

\noindent {\sf onekind(heard).} \hspace{.2em} {\sf allkind(like).} \hspace{.2em} 
{\sf all\_onekind(own).} \hspace{.2em} {\sf propkind(alpha).}

(A more extensive use of the list notation would allow to write rules  dealing with any arity in a yet more natural way.)\\

Writing {\sf onekind(heard)} means that the classes or objects
which can appear as parameters of the unary predicate $heard$ of the formal 
system have adapted meaning. In this example, $bell$ may denote the class
of some bells considered in the situation at hand, $loud\_bell$ the class of those bells which are loud and $some\_bell$ could denote a precise $loud\_bell$.
Notice that the objects are intended to denote either individuals or classes of 
individuals.

In this situation the user could add the following atom to the one 
given above about bells: {\sf ont\_object(some\_bell,loud\_bell).}

As an example, let us consider our binary predicate  $own$,
with the known informations provided by the user as follows in ASP:

\noindent{\sf ont\_object(tom,student).} \hspace{1em}
{\sf ont\_object(book,document).}

Then, the following ontological atoms would be deduced by the system
(again in ASP notation):

\noindent{\sf ont([own,tom,book], [own,tom,document])}\\
{\sf ont([own,student,book], [own,tom,book])}\\
{\sf ont([own,student,book], [own,tom,document])}.\\

A predicate for which no ``all/one'' information is given  could be used, but it 
would never give rise to ontological atoms.

This kind of formulation is in accordance with our requirements for the formalism:
allowing serious expressiveness while keeping computation manageable.
To this respect, for large examples it can be useful to reduce the instantiation size by
restricting the parameters which can be used for some predicates,
by adding {\sf not unrestr(P), kindPar(P,X,Y)} to the body of the rule 
{\tt(O-O)}. Rules such as the following ones should be added\\
{\sf unrestr(P) :- not restr(P), pred(P).}
{\sf pred(P) :- onekind(P).} , [...{\sf allkind(P)},...].\\ 
The user would provide the relevant information about 
{\sf restr} and {\sf kindPar}.

\subsection{Conclusion and future work}
We have shown how a recent version of running ASP systems allow 
easy translation of logical formalisms. The example given here is a formalism allowing to infer ``explanations" from ``causal and ontological" information, with two requirements:
1 It must be easy and natural to formalize a given situation.
2 Computation should remain as manageable as possible.

These requirements justify  some restrictions of our formalism, in particular the
fact that we accept 
only classical atoms inside the causal and ontological atoms, without operators such as negation or disjunction.
The formalism is not restricted to the propositional case, and,
as explained in the last subsection, it involves an ontology slightly more general that the rudimentary one developed in the full ASP description given in 
the preceding subsections.

From the perspective of existing (and predictable in a near future) ASP systems,
here are some conclusions that can be drawn:

The existing recent extensions of ASP systems are a real bonus.
It is much easier to {\em write} the program now than it was a few years ago.
More importantly perhaps, the difference is even greater when {\em reading} such a
program.
Indeed the claimed assertions about ASP, namely that it makes programs easy to write and to read, are true only when such extensions are used.
Otherwise, due to the fact that in practice it is very hard to use some portion of a program in different places in a greater program, the real ASP programs are either very restricted in their application, or hard to read (and thus to evolve).
In our opinion, DLV with templates (DLT) solves a serious part of this problem.
However, on our formalism, it happens that it is DLV-Complex, since it 
accepts sets and lists in a natural and efficient way, which has made the difference.
Indeed, our formalism involves sets in a non trivial way, thus the present program is far more readable than the previous ones, in pure DLV. 

From a formal perspective, neither DLV-Complex nor DLT are revolutionary. To a great extent, all they do is add way for writing natural programs where pure ASP systems need cumbersome rewritings of parts of the programs. Generally, using them does not improve computational efficiency.
The fact is that they always allow more convenient writing, and particularly 
much easier reading, which changes the life of the programmer.
Expert ASP programmers were able to use previous ASP systems together with 
small parts of more standard programming, but this was not very attracting for most of the potential users of ASP systems.

So, what can we expect now?

For now, DLT cannot work together with DLV-Complex, and this is the 
first thing we can expect, hopefully in a very near future, together with 
similar facilities for gringo/clasp and other systems.

Also, ``brave'' and ``cautious'' reasoning do not exist in full generality (to 
our knowledge) in available  systems. May be our formalism is an extreme example of this point (a mix of cautious and standard reasoning would be useful), 
but in fact 
this happens to be a serious concern in many circumstances.
Again, programmers can redirect the result (the set of all answer sets) 
to a very simple other ASP program, for playing with the answer sets in a 
more or less complex way, but this is not very convenient.
In the literature, we can encounter various texts about this problem, 
describing systems which solve it in a natural manner, but, to our knowledge, 
such systems are not yet available.
Some [meta?]predicates, as described in e.g. \cite{TBA05}, 
which present a relatively simple and natural way to write programs doing this,
would suffice in many cases. 
In this way real brave and cautious reasoning could be envisioned, and much more.
We hope that this will be the case in available systems in a near future.
Clearly, dealing with too many answer sets cannot be done with huge domains, but
there are many cases where it would be interesting to have full access to the 
set of all the answer sets.

Here is a last feature that could be interesting.
The full program did not work 
for what is called our ``great example''. However, when split in two or three parts, it worked relatively well. In this case the split was very easy to do.
Could it be that future systems detect such possible split, and thus extend their range of application?
In our example, computing {\sf ecSet} first, then {\sf ecSetRes} and finally {\sf ecSetVer} should be automatized. It seems easy to detect such one way dependencies, without retro-action. The great difference in practice between 
launching the three programs together, and launching them one after the other,
shows that such improvement could have  spectacular consequences.
 
For what concerns our own work, the important things to do are to apply the formalism to real situations, and, to this respect, firstly to significantly extend our notion of ``ontology'' towards a real one.

\section*{Acknowledgement} 
The author thanks the reviewers for their helpful and constructive comments,
and Marie-Odile Cordier and Philippe Besnard who made this work possible.

\end{document}